\documentclass[preprint,12pt]{elsarticle}

\usepackage{graphicx}
\usepackage{amsmath,amssymb,amsfonts}
\usepackage{algorithm}
\usepackage{algorithmicx}
\usepackage[noend]{algpseudocode}
\usepackage{multirow}
\usepackage{xcolor}
\usepackage{breqn}
\usepackage{url}
\usepackage{ulem}
\usepackage{cleveref}
\usepackage{appendix}
\usepackage{url}
\newcommand{\Bk}{\color{black}}

\newcommand{\bftheta}{\boldsymbol{\theta}}

\setcitestyle{square}

\DeclareMathOperator*{\argmin}{arg\,min}

\journal{Computers and Mathematics with Applications}

\begin{document}

\begin{frontmatter}



\title{Solutions to Elliptic and Parabolic Problems via Finite Difference Based Unsupervised Small Linear Convolutional Neural Networks}


\affiliation[rice]{organization={Rice University},
            city={Houston},
            postcode={77005}, 
            state={TX},
            country={USA}}

\affiliation[mda]{organization={The University of Texas MD Anderson Cancer Center},
            city={Houston},
            postcode={77003}, 
            state={TX},
            country={USA}}
            
\author[rice,mda]{Adrian Celaya}
\author[rice]{Keegan Kirk}
\author[mda]{David Fuentes}
\author[rice]{Beatrice Riviere}

\begin{abstract}
In recent years, there has been a growing interest in leveraging deep learning and neural networks to address scientific problems, particularly in solving partial differential equations (PDEs). However, many neural network-based methods like PINNs rely on auto differentiation and sampling collocation points, leading to a lack of interpretability and lower accuracy than traditional numerical methods. As a result, we propose a fully unsupervised approach, requiring no training data, to estimate finite difference solutions for PDEs directly via small linear convolutional neural networks. Our proposed approach uses substantially fewer parameters than similar finite difference-based approaches while also demonstrating comparable accuracy to the true solution for several selected elliptic and parabolic problems compared to the finite difference method.
\end{abstract}


\end{frontmatter}


\section{Introduction}
\label{sec:intro}
Partial differential equations (PDEs) of elliptic and parabolic type are ubiquitous in the mathematical modeling of many physical phenomena and thus see wide application to many real-world problems. Classical numerical methods (i.e., finite difference and element methods) introduce a computational mesh over which we define representations of differential operators with matrices. This results in large linear systems whose efficient solution presents several computational challenges. Physics informed neural networks (PINNs) have recently gained popularity in solving PDEs \cite{Hornik, Funahashi, dissanayake1994neural, lagaris1998artificial, raissi2017physics, raissi2017physics2}. This approach uses the universal approximation property of deep neural networks to train a surrogate model (i.e., the neural network) of the solution to a given PDE. 

A vital aspect of PINNs is using auto differentiation to compute a residual-based loss function for a set of sampled collocation points \cite{hou2023enhancing}. While PINNs represent a mesh-free solution and have shown promise in multiple fields like biology \cite{zhang2023physics, sarabian2022physics}, meteorology \cite{li2023lpt}, and optimal control \cite{mowlavi2023optimal, fotiadis2023physics}, this reliance on auto differentiation and sampling results in a lack of interpretability and lower accuracy than traditional numerical methods \cite{chiu2022can}. Choosing collocation points involves sampling random points from a uniform distribution on the computational domain, and it is unclear how many points are required to achieve acceptable results for a given PDE \cite{chiu2022can, hou2023enhancing, tang2023pinns}. Additionally, while networks can express very complex functions, determining the appropriate architecture and number of parameters to solve a specific PDE can be challenging \cite{ramachandran2017searching, cuomo2022scientific, karniadakis2021physics}. If a specific PDE requires a neural network with many parameters, it will increase the memory and time costs required for training. Finally, the generalization capability of PINNs to points outside their training domains is unclear, and a topic of ongoing research \cite{bonfanti2023generalization}.  

\begin{figure}
    \centering
    \includegraphics[width=\textwidth]{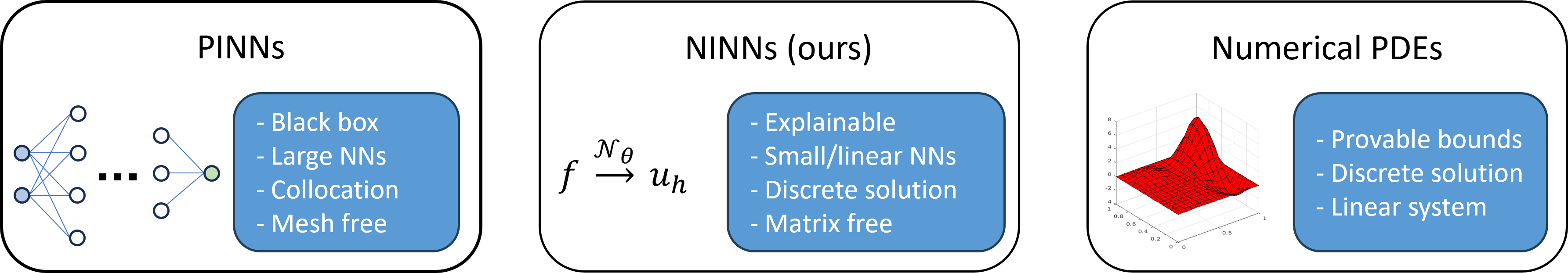}
    \caption{Properties of classical PINNs, numerics informed neural networks (NINNs), and numerical PDEs.}
    \label{fig:intro-figure}
\end{figure}

While traditional numerical methods (such as finite difference and finite element methods) are computationally expensive, they are highly structured and can guide the design of neural network-based methods so that they output explainable solutions and are more computationally efficient. If we constrain a neural network-based approach to mirror a numerical method, it is reasonable to expect solutions from the neural network to be close to those of a numerical method with known error bounds. Additionally, the structure of numerical methods makes it possible to get away with using simpler neural networks that converge to small loss values. Examples of such neural network-based methods include the Deep Ritz and Deep Galerkin methods \cite{SirignanoDGM:2018, WangDRM:2018,MullerZeinhofer}. Both of these methods incorporate aspects of finite element methods, allowing for faster convergence during training by utilizing the structure and properties of the numerical method that inspired them. 

This work proposes a finite difference-based approach to approximating solutions to elliptic and parabolic PDEs that utilizes small, linear convolutional neural networks (CNNs). Figure~\ref{fig:intro-figure} displays the properties of our proposed method which we call a ``numerics informed neural network'' (NINN), classical PINNs, and numerical PDEs. Our choice of architecture, inspired by the PocketNet paradigm \cite{celaya2022pocketnet}, uses CNNs that closely mirror geometric multigrid methods to almost exactly recover finite difference solutions with fewer parameters than other CNN-based methods. To the best of our knowledge, we are also the first to propose using linear CNNs, which do not use non-linear activation functions, to learn finite difference solutions. 

Finite difference methods have inspired the construction of various neural networks \cite{chiu2022can, Lim2022}, but these methods require training data. The recent work \cite{Zhao2023} solves elliptic PDEs with constant diffusion, and it shares the following similarities with our proposed numerical method: no training data is needed, and the loss function is derived from the five-point stencil of the finite difference approximation. However, our loss function is different as it incorporates the Dirichlet boundary condition in a weighted fashion, whereas \cite{Zhao2023} directly enforces the boundary condition outside of the loss function (more details in Section~\ref{sec:results}). We also define the method for elliptic PDEs with non-constant diffusion coefficients and extend it to time-dependent problems. Another key feature of our method is using small neural networks, which makes it computationally efficient.

An outline of the paper is as follows. Section~\ref{sec:methods} introduces the model problem, the network architecture and the loss function.   Section~\ref{sec:calculation} describes the algorithms for both steady-state and time-dependent problems. Results and discussion are presented in Section~\ref{sec:results}. Conclusions follow. 

\section{Material and Methods} \label{sec:methods}
We propose a fully unsupervised method for estimating finite differences solutions to partial differential equations via convolutional neural networks. In contrast to existing deep learning-based methods, our approach is fully unsupervised. In other words, our approach does not require training data and estimates the solution to a given PDE directly via the optimization process, also called the training process. 

\subsection{Elliptic Problems}
Let $\Omega \subset \mathbb{R}^2$ be an open set. Given a function $f:\Omega \rightarrow \mathbb{R}$ and diffusion coefficients $\kappa$, we consider the 2D Poisson problem with solution $u: \overline{\Omega} \rightarrow \mathbb{R}$ such that
\begin{align} \label{eqn:elliptic}
    \begin{split}
    -\nabla \cdot (\kappa \nabla u) &= f \text{ in } \Omega, \\
    u &= g \text{ on } \partial \Omega.
    \end{split}
\end{align}
We begin with the case when $\kappa$ is constant and assume that $\kappa = 1$. For readibility, we assume that $\Omega$ is the square domain $(0,L)^2$, that is partitioned into a uniform grid of $N\times N$ squares with size $h=L/N$. The methodology presented in the paper can be easily extended to rectangular domains.  We first recall the standard finite difference method based on the five-point stencil, applied to \eqref{eqn:elliptic}. Let $\mathcal{V}^0$ (resp. $\mathcal{V}^\partial$) be the set of indices $(i,j)$ such that the point $(x_i,y_j) = (i h, j h)$ belongs to the interior (resp. boundary) of $\Omega$.  
\begin{align} \label{eqn:fd-scheme}
    \frac{4u_{i,j} - u_{i+1, j} - u_{i-1, j} - u_{i, j+1} - u_{i, j-1}}{h^2} = f(x_i,y_j), \quad \forall (i,j) \in \mathcal{V}^0, \\
    u_{i,j} = g(x_i,y_j), \quad \forall (i,i)\in\mathcal{V}^\partial. \label{eqn:fd-schemeb}
\end{align}
The finite difference solution $u_h$ is a vector with entries $u_{i,j}$ satisfying \eqref{eqn:fd-scheme}-\eqref{eqn:fd-schemeb} and it is known that for smooth enough solutions $u$ and small enough $h$, the value $u_{i,j}$ is a good approximation of $u(x_i,y_j)$. One can rewrite \eqref{eqn:fd-scheme} in terms of a discrete convolutional operator $\star$ and convolutional kernel $K_\Delta$ defined as:
\begin{gather}
    K_{\Delta} = \frac{1}{h^2}\begin{bmatrix} 0 & -1 & 0 \\ -1 & 4 & -1 \\ 0 & -1 & 0\end{bmatrix}.
\end{gather}
We use the shorthand notation $f_{i,j} = f(x_i,y_j)$. 
The equivalent form for \eqref{eqn:fd-scheme} is
\begin{align} \label{eqn:fd-conv}
    (K_{\Delta} \star u_h)_{i,j} = f_{i,j}, \quad \forall (i,j)\in\mathcal{V}^0.
\end{align}
For a given kernel $K \in \mathbb{R}^{3 \times 3}$, the convolution operator $\star$ is defined by
\begin{align}
    (K \star u_h)_{i,j} = \sum_{p=-1}^1 \sum_{q=-1}^1 K_{p,q} \, u_{i+p,j+q}.
\end{align}
With this convolution-based discretization, we can now reformulate the finite difference approximation of (\ref{eqn:elliptic}) as a convex optimization problem given by
\begin{dmath} \label{eqn:opt}
    \argmin_{u_h} \sum_{(i,j) \in \mathcal{V}^0} \left(\left(K_{\Delta} \star u_h\right)_{i,j} -f_{i,j}\right)^2 + \sum_{(i,j) \in \mathcal{V}^\partial} \left(\left( u_h \right)_{i,j} - g_{i,j}\right)^2.
\end{dmath}
We obtain an approximate solution $\hat{u}$ to \eqref{eqn:opt} by training a neural network $\mathcal{N}_{\theta}: \mathbb{R}^{N \times N} \rightarrow \mathbb{R}^{N \times N}$ with trainable parameters $\bftheta\in\mathbb{R}^M$.  We propose to train $\mathcal{N}_{\theta}$ using the unsupervised loss function $\mathcal{L}_{\alpha}: \mathbb{R}^{N \times N} \rightarrow \mathbb{R}$ given by
\begin{align}
\label{eqn:loss}
    \mathcal{L}_{\alpha}(\hat{u}) = \alpha \sum_{(i,j) \in \mathcal{V}^0} \left(\left(K_{\Delta} \star 
    \hat{u}\right)_{i,j} - f_{i,j}\right)^2 + (1 - \alpha) \sum_{(i,j) \in \mathcal{V}^\partial} \left(\hat{u}_{i,j} - g_{i,j}\right)^2,
\end{align}
where $\alpha = h^2 / 4$ is a weighting term. The idea behind this loss function is to directly estimate the finite difference approximation to (\ref{eqn:elliptic}) without using any training data. Specifically, we use $f$ as the input to the CNN and we perform $\mathcal{M}$ updates of the parameters in the neural network $\mathcal{N}_\bftheta$, using the loss function defined above.   Algorithm \ref{alg:alg-elliptic} in Section \ref{sec:calculation} outlines this training procedure in more detail.

In the case where $\kappa$ is piecewise constant, $\kappa = (\kappa_{i,j})$, we use the following finite difference discretization
\begin{multline} \label{eqn:non-const-fd}
    -\frac{1}{h^2}\left( \kappa_{i+\frac{1}{2}, j}(u_{i+1, j} - u_{i,j}) - \kappa_{i-\frac{1}{2}, j}(u_{i, j} - u_{i-1,j}) \right) \\
    -\frac{1}{h^2}\left( \kappa_{i, j+\frac{1}{2}}(u_{i, j+1} - u_{i,j}) - \kappa_{i, j-\frac{1}{2}}(u_{i, j} - u_{i,j-1}) \right)
    = f_{i, j},
\end{multline}
where the interface values $\kappa_{i+\frac{1}{2}, j}$, $\kappa_{i-\frac{1}{2}, j}$, $\kappa_{i, j+\frac{1}{2}}$, and $\kappa_{i, j-\frac{1}{2}}$ are the harmonic averages of $\kappa_{i,j}$ given by
\begin{align}
    \begin{split}
        \kappa_{i+\frac{1}{2}, j} &= \frac{2}{\frac{1}{\kappa_{i + 1, j}} + \frac{1}{\kappa_{i,  j}}} \hspace{18mm} 
        \kappa_{i-\frac{1}{2}, j} = \frac{2}{\frac{1}{\kappa_{i, j}} + \frac{1}{\kappa_{i - 1,  j}}} \\
        \kappa_{i, j+\frac{1}{2}} &= \frac{2}{\frac{1}{\kappa_{i, j + 1}} + \frac{1}{\kappa_{i,  j}}} \hspace{18mm} 
        \kappa_{i, j-\frac{1}{2}} = \frac{2}{\frac{1}{\kappa_{i, j}} + \frac{1}{\kappa_{i,  j - 1}}}.
    \end{split}
\end{align}

The interface values necessitate the imposition of a half-grid (also called dual grid) onto our original grid. As a result, we modify our neural network prediction via a dilation operator $D:\mathbb{R}^{N \times N} \rightarrow \mathbb{R}^{(2N - 1) \times (2N - 1)}$ such that
\begin{align}
    D(\hat{u})_{i, j} = \begin{cases}
        \hat{u}_{1 + \frac{i-1}{2}, 1 + \frac{j-1}{2}}, & \text{ if } i \text{ and } j \text{ are odd} \\
        0, &\text{otherwise.}
    \end{cases}
\end{align}
Additionally, we map $\kappa$ from the original grid to the dual grid by applying the following steps:\\
Step 1: apply the dilation operator to obtain $\kappa^\ast\in\mathbb{R}^{(2N-1)\times(2N-1)}$:
\begin{align}
\kappa^\ast = D(\kappa).
\end{align}
Step 2: update the values of $\kappa^\ast$ on the half-grid:
\begin{align}
\kappa^{\ast}_{i,j} =  \begin{cases} 2\left(\frac{1}{\kappa^\ast_{i-1,j}} + \frac{1}{\kappa^\ast_{i+1, j}}\right)^{-1},  & \text{ if } i \text{ odd and  } j \text{ even} \\
    2\left(\frac{1}{\kappa^\ast_{i,j-1}} + \frac{1}{\kappa^\ast_{i, j+1}}\right)^{-1},  & \text{if } i \text{ even and } j \text{ odd} 
    \end{cases} 
\end{align}
Step 3: set to zero the values of $\kappa^\ast$ that would correspond to the values on the original grid:
\begin{align}
\tilde{\kappa} = \left( \mathbf{1}_{(2N - 1) \times (2N - 1)} - D(\mathbf{1}_{N \times N}) \right) \odot \kappa^\ast.
\end{align}
In the equation above, $\mathbf{1}_{N \times N}$ is an $N \times N$ matrix with all entries equal to one,  and $\odot$ is the Hadamard or pointwise product.

We now introduce the following convolution kernels
\begin{align}
    \begin{split}
    T_{\uparrow} &= \begin{bmatrix} 0 & 0 & 1 & 0 & 0 \\ 0 & 0 & 0 & 0 & 0 \\ 0 & 0 & -1 & 0 & 0 \\ 0 & 0 & 0 & 0 & 0 \\ 0 & 0 & 0 & 0 & 0 \end{bmatrix} \hspace{3mm} 
    P_{\uparrow} = \begin{bmatrix} 0 & 0 & 0 & 0 & 0 \\ 0 & 0 & 1 & 0 & 0 \\ 0 & 0 & 0 & 0 & 0 \\ 0 & 0 & 0 & 0 & 0 \\ 0 & 0 & 0 & 0 & 0 \end{bmatrix} \\
    T_{\leftarrow} &= -T_{\uparrow}^T \hspace{26.95mm} P_{\leftarrow} = P_{\uparrow}^T \\
    T_{\downarrow} &= \begin{bmatrix} 0 & 0 & 0 & 0 & 0 \\ 0 & 0 & 0 & 0 & 0 \\ 0 & 0 & 1 & 0 & 0 \\ 0 & 0 & 0 & 0 & 0 \\ 0 & 0 & -1 & 0 & 0 \end{bmatrix} \hspace{3mm} 
    P_{\downarrow} = \begin{bmatrix} 0 & 0 & 0 & 0 & 0 \\ 0 & 0 & 0 & 0 & 0 \\ 0 & 0 & 0 & 0 & 0 \\ 0 & 0 & 1 & 0 & 0 \\ 0 & 0 & 0 & 0 & 0 \end{bmatrix} \\
    T_{\rightarrow} &= -T_{\downarrow}^T \hspace{26.95mm} P_{\rightarrow} = P_{\downarrow}^T, \\
    \end{split}
\end{align}
which will allow us to rewrite \eqref{eqn:non-const-fd} as follows:
\begin{multline} \label{eqn:non-const-est}
    -\frac{1}{h^2} \left( \left(T_{\uparrow} \star D(u)\right) \odot \left(P_{\uparrow} \star \tilde{\kappa}\right) - \left(T_{\downarrow} \star D(u)\right) \odot \left(P_{\downarrow} \star \tilde{\kappa}\right) \right.\\
    \left. + \left(T_{\rightarrow} \star D(u)\right) \odot \left(P_{\rightarrow} \star \tilde{\kappa}\right) - \left(T_{\leftarrow} \star D(u)\right) \odot \left(P_{\leftarrow} \star \tilde{\kappa}\right) \right)_{i,j} = f_{i,j}, \forall i,j \in \mathcal{V}^0, \quad
\end{multline}
which is then used in the first term of the loss function \eqref{eqn:loss} to train our neural network. The same training process outlined in Algorithm \ref{alg:alg-elliptic} applies here. Note that in \eqref{eqn:non-const-est}, the convolution operator $\star$ uses a stride of two because of the dual grid and the use of larger convolution kernels.

\subsection{Parabolic Problems}
We generalize the method described in the previous section to time-dependent problems. We propose a modified loss function that incorporates the backward Euler method to address the time derivative. Consider the following parabolic problem on a square domain $\Omega$ and for the time interval $[t_0,T]$. 
\begin{align}
\label{eqn:parabolic}
    \begin{split}
    \frac{\partial u}{\partial t} - \Delta u &= f \quad \mbox{in}\quad  \Omega \times (t_0, T], \\
    u &= g \quad\mbox{on}\quad \partial \Omega \times (t_0, T], \\
    u(\cdot,t_0)  &= u^0 \quad\mbox{in}\quad \Omega\times \{t_0\}.
    \end{split}
\end{align}
Let $\tau>0$ be the time step size. At each time step, $t^n = n\tau$, the finite difference solution denoted by $u_h^n$ solves 
the following optimization problem 
\begin{align} \label{eqn:timeopt}
    \argmin_{u_h^n} & \sum_{(i,j) \in \mathcal{V}^0} \left(u^n_{i,j} - u^{n-1}_{i,j} - \tau\left(f_{i,j}^n - \left(K_{\Delta} \star u_h^n\right)_{i,j}\right)\right)^2 \nonumber\\
    &+ \sum_{(i,j) \in \mathcal{V}^\partial} \left(u^n_{i,j} - g_{i,j}^n\right)^2.
\end{align}
Above we use the short-hand notation $f_{i,j}^n = f(x_i,y_j,t^n)$ and $g_{i,j}^n = g(x_i,y_j,t^n)$. Here, the unknowns $u_{i,j}^n$ are the approximations of the solution $u$ evaluated at the point $(x_i, y_j)$ and time $t^n$. We then use the following loss function to train our CNN at each time step
\begin{align} \label{eqn:td-loss}
    \mathcal{L}_{\alpha, \tau}(\hat{u}^n, \hat{u}^{n-1}, t^n) = & \alpha \sum_{(i,j) \in \mathcal{V}^0} \left(\hat{u}^n_{i,j} - \hat{u}^{n-1}_{i,j} - \tau\left(f_{i,j}^n - \left(K_{\Delta} \star \hat{u}^n\right)_{i,j}\right)\right)^2  \nonumber\\
    & + (1 - \alpha) \sum_{(i,j) \in \mathcal{V}^\partial} \left(\hat{u}^n_{i,j} - g_{i,j}^n\right)^2.
\end{align}
Again, we will choose the weight $\alpha = h^2 / 4$. The key difference between \eqref{eqn:loss} and \eqref{eqn:td-loss} is that we minimize at each time step. The input to the CNN at each time step is the previous solution $\hat{u}^{n-1}$ and the output is the solution at the time step $t^n$. Algorithm \ref{alg:alg-parabolic} in Section \ref{sec:calculation} outlines this training procedure in more detail.

\subsection{Network Architecture} \label{sec:architecture}
\begin{figure}[ht!]
    \centering
    \includegraphics[width=\textwidth]{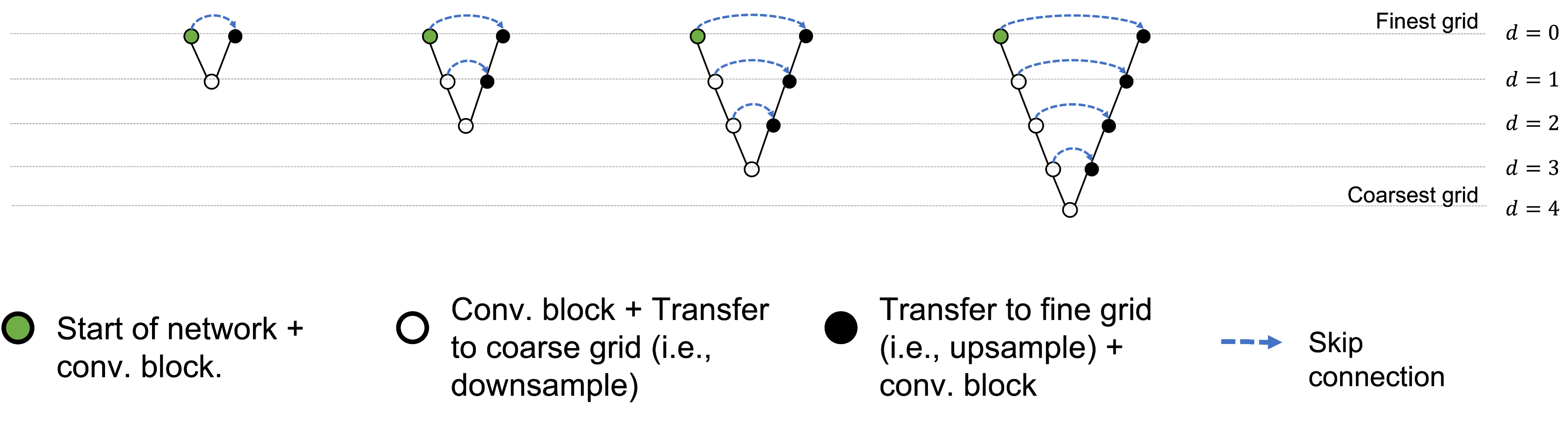}
    \caption{Sketch of U-Net architecture for different network depths: $0\leq d\leq 4$. \label{fig:network}}
\end{figure}

We select as our architecture the widely used U-Net \cite{unet}. The U-Net was initially designed for image segmentation tasks but is successful in several scientific machine learning tasks \cite{gravity1, gravity2, joint-inversion}. The architecture consists of convolution block operators (blocks), a downsampling (or encoding) path, and an upsampling (or decoding) path. A channel-wise concatenation operation links each layer in the downsampling and upsampling paths. Each block consists of two 2D convolution layers.  Unlike most implementations of the U-Net architecture \cite{unet, unet-3d, nnunet}, we omit normalization (i.e., batch or instance normalization) and do not apply a non-linear activation function like ReLU and use the identity function instead (see Section \ref{sec:results}). Additionally, we use the PocketNet approach proposed in \cite{celaya2022pocketnet} and leave the number of feature maps (channels) at each resolution within our architecture constant, namely $32$.  Figure \ref{fig:network} sketches the U-Net architecture for different network depths. We define the depth, $d$, of each network to be the number of downsampling operations present in the network. The size of the feature tensor is halved from depth $d$ to depth $d+1$. 

\subsection{Data} \label{sec:data}

\subsubsection{Elliptic Problem Data} \label{sec:data-elliptic}
We test our proposed approach for elliptic problems on four different test cases.  Unless otherwise specified, the domain is the unit square.

\paragraph{The Bubble Function} In the first test case, (\ref{eqn:elliptic}) is solved with data chosen such that the exact solution is the bubble function 
\begin{align}
    u(x, y) = x(x-1)y(y-1).
\end{align} 
This function is symmetric, and we apply homogeneous zero Dirichlet boundary conditions. Note that for this function, one can show that the five-point stencil scheme given by (\ref{eqn:fd-scheme}) applied to the bubble function gives exactly $-\Delta u$ evaluated at the grid point $(x_i, y_j)$, which implies that the finite difference approximation is exact.
\paragraph{The ``Peak'' Function} In the second case, we want to test the effectiveness of our strategy when the solution has high gradients and is non-symmetric. Hence, we select data such that the exact solution is the ``peak'' function
\begin{align}
    u(x, y) = 0.0005\left( x(x-1)y(y-1) \right)^2 e^{10x^2 + 10y}.
\end{align}
\paragraph{The Exponential-Trigonometric Function} We test our approach on a function with non-homogeneous Dirichlet boundary conditions in the third case. We select data such that the exact solution is the exponential-trigonometric function given by
\begin{align}
    u(x, y) = e^{-x^2 - y^2}\sin(3 \pi x)\sin(3 \pi y) + x.
\end{align}
\paragraph{Non-Constant and Discontinuous Diffusion} 
To test how our approach fairs on problems with low regularity, we consider an example with a non-constant, discontinuous diffusion coefficient taken from \cite{riviere2003posteriori}. The problem is posed on the square domain $\Omega = (-1,1)^2$ which we divide into four subdomains $\Omega_i$ corresponding to the quadrants of the Cartesian plane. On each subdomain, $\kappa$ is constant and takes the values $\kappa_1 = \kappa_3 = 5$ and $\kappa_2 = \kappa_4 = 1$. 
The exact solution takes the form
\begin{align}
r^{\beta}(a_i \sin(\beta \theta) + b_i \cos(\beta \theta)),
\end{align}
where $(r, \theta)$ are the polar coordinates of a given point in $\Omega$, $a_i, b_i$ are constants that depend on the subdomains (see \cite{riviere2003posteriori} for exact values). This discontinuity in the diffusion coefficient introduces a singularity in the solution at the origin, namely the function belongs to the Sobolev space $H^{1+\beta}(\Omega)$, with $\beta \approx 0.5354$.


\subsubsection{Parabolic Problem Data} \label{sec:data-parabolic}
We test our proposed approach for parabolic problems on three different test cases.
\paragraph{The Trigonometric Function} In the first two cases, (\ref{eqn:parabolic}) is solved with data such that the exact solution is given by the trigonometric function
\begin{align}
    u(x, y, t) = \cos(t)\sin(n \pi x)\sin(n \pi y).
\end{align}
In the first case, we select $n=1$, which gives a symmetric function with a single peak. In the second case, we set $n=4$, which results in 16 peaks and troughs in our domain. 

\paragraph{The Gaussian Function} In the third case, we test our approach on a  Gaussian function centered on the point $\left(\frac{1}{2}, \frac{1}{2}\right)$
\begin{align}
    u(x, y, t) = \cos(t)e^{-50\left((2x - 1)^2 + (2y - 1)^2\right)}.
\end{align}

\subsection{Training and Testing Protocols}
 We use the Adam optimizer \cite{adam} and 5$\times$5 convolutional kernels in each layer. The initial learning rate is 0.001 for the steady-state elliptic problems and 0.0001 for parabolic problems. During training, we use L2 regularization with a penalty of $10^{-7}$ and the norm of the network's gradient is clipped so that it is no greater than $10^{-2}$. Our models are implemented in Python using TensorFlow (v2.12.0) and trained on an NVIDIA Quadro RTX 6000 GPU \cite{keras}. All network weights are initialized using the default TensorFlow initializers. All other hyperparameters are left at their default values. The code for our network architecture is available at \url{https://github.com/aecelaya/pde-nets}.

To assess the accuracy of our predicted solutions, we use the following norms of the error between the exact solution $u$ and its approximate $\hat{u}$ for steady-state problems:
\begin{align}
    ||u - \hat{u}||_{2,h} &= h \sqrt{\sum_{(i,j)\in\mathcal{V}^0\cup\mathcal{V}^\partial} (u(x_i,y_j) - \hat{u}_{i,j})^2} \\
    ||u - \hat{u}||_{\infty} &= \max_{(i,j)\in\mathcal{V}^0\cup\mathcal{V}^\partial} |u(x_i,y_j) - \hat{u}_{i,j}|.
\end{align}
For time-dependent problems, similar errors are computed at the final time $T$.

\section{Calculation} \label{sec:calculation}
We use Algorithm \ref{alg:alg-elliptic} to approximate a solution $\hat{u}$ to (\ref{eqn:elliptic}). In this algorithm, we start by initializing the neural network $\mathcal{N}_{\theta}$ with trainable parameters $\bftheta$. In this case, we use the U-Net architecture described in Section \ref{sec:architecture}. We set the maximum number of optimization steps $\mathcal{M}$ as the stopping criterion for the algorithm. We then begin the minimization (or training) process by generating a prediction from our neural network with the source term $f$ as the input. Given the current prediction, we compute the loss value and update the network weights via backpropagation. If the loss value from the current prediction is less than the previous best value, we update our best loss and save the current prediction. Algorithm \ref{alg:alg-elliptic} fully outlines this procedure.

\begin{algorithm}[ht!]
\caption{Unsupervised CNN Training For (\ref{eqn:elliptic}) \label{alg:alg-elliptic}}
 \hspace*{\algorithmicindent} \textbf{Input} Right-hand side $f$ and boundary condition $g$ \\
 \hspace*{\algorithmicindent} \textbf{Output} Approximate solution to (\ref{eqn:opt}), $\hat{u}_*$
\begin{algorithmic}[1]
\State Randomly initialize the network parameters $\bftheta$
\State Set maximum iterations $\mathcal{M}$
\State $k \gets 0$ \Comment{Iteration counter}
\State $\ell_* \gets +\infty$ \Comment{Store smallest loss value}
\While{$k < \mathcal{M}$}
\State $\hat{u} \gets \mathcal{N}_{\theta}(f)$ \Comment{Get network prediction}
\State $\ell \gets \mathcal{L}_{\alpha}(\hat{u})$ \Comment{Compute loss}
\State Update $\bftheta$ using backpropagation on $\ell$
\If{$\ell < \ell_*$}
\State $\hat{u}_* \gets \hat{u}$ \Comment{Save best prediction}
\State $\ell_* \gets \ell$ \Comment{Update smallest loss value}
\EndIf
\State $k \gets k+1$
\EndWhile
\end{algorithmic}
\end{algorithm}


We use Algorithm \ref{alg:alg-parabolic} to approximate a solution $\hat{u}^{N_T}$ to (\ref{eqn:parabolic}), where $N_T$ is an integer such that $T = N_T \tau$. Like with Algorithm \ref{alg:alg-elliptic}, we randomly initialize a U-Net and set a maximum number of optimization steps as a stopping criterion. However, in this case, the maximum number of steps applies to each time step. We then set the initial condition as the previous solution. We apply nearly the same steps for each time step as with Algorithm \ref{alg:alg-elliptic}. Namely, we produce a candidate prediction from our neural network with the previous solution as the input. We compute the loss value and update the network weights via backpropogation on the loss. If the loss value is lower than the previously observed lowest loss value, then we update our lowest loss value and save the current prediction. At the end of this process, we set the previous solution to the best current solution. This process continues for every time step. Algorithm \ref{alg:alg-parabolic} fully outlines this procedure. Note that in this algorithm, we do not reinitialize our neural network weights at each time step. Instead, we use the previous network configuration as the initial state for the next time step.

\begin{algorithm}[ht!]
\caption{Unsupervised CNN Training For (\ref{eqn:parabolic}) \label{alg:alg-parabolic}}
 \hspace*{\algorithmicindent} \textbf{Input} Number of time steps $N_T$, final time $T$, initial condition $u^0$, \\
  \hspace*{\algorithmicindent} right hand side $f$, and boundary condition $g$ \\
 \hspace*{\algorithmicindent} \textbf{Output} Approximate solution to (\ref{eqn:timeopt}) at final time $\hat{u}^{N_T}$
\begin{algorithmic}[1]

\State Randomly initialize the network parameters $\bftheta$
\State Set maximum iterations $\mathcal{M}$
\State $\tau \gets T/N_T$ \Comment{Initialize time step size}
\State $n \gets 1$ \Comment{Time step counter}
\While{$n \leq N_T$}

\State $k \gets 0$ \Comment{Iteration counter}
\State $\ell_* \gets +\infty$ \Comment{Store smallest loss value}
\While{$k < \mathcal{M}$}
\State $w \gets \mathcal{N}_{\theta}(\hat{u}^{n-1})$ \Comment{Get network prediction}
\State $\ell \gets \mathcal{L}_{\alpha, \tau}(w, \hat{u}^{n-1}, n \tau)$ \Comment{Compute loss}
\State Update $\bftheta$ using backpropagation on $\ell$
\If{$\ell < \ell_*$}
\State $\hat{u}^n \gets w$ \Comment{Save best prediction}
\State $\ell_* \gets \ell$ \Comment{Update smallest loss value}
\EndIf
\State $k \gets k+1$
\EndWhile
\State $n \gets n + 1$
\EndWhile
\end{algorithmic}
\end{algorithm}
\Bk

\section{Results and Discussion}\label{sec:results}
Using the methods described in Section \ref{sec:methods} and Algorithm \ref{alg:alg-elliptic}, we approximate the finite difference solution to the elliptic problems defined in Section \ref{sec:data-elliptic} with constant diffusion coefficients (i.e., $\kappa=1$). Table \ref{tab:results-elliptic-grid-opt-steps} shows the accuracy of our unsupervised predictions for a varying grid size and number of optimization steps. We fix the depth of our U-Net architecture to three in this case. For reference, Table \ref{tab:fd-results} shows the accuracy of the finite difference method for varying grid sizes. For the peak and exponential-trigonometric functions, our approach almost exactly recovers the finite difference solution. We do not, however, see this for the bubble function example. In that case, the solution of the finite difference method is exact. Under our selected settings (i.e., depth and optimization steps), our method stops at an error of approximately 10$^{-6}$ in the $|| \cdot ||_{2, h}$ norm for the bubble function example. The cause of this discrepancy between our method applied to the bubble function and the finite difference solution is due mostly to the fact that single-precision is used to train the neural networks.  Values of the order $10^{-6}$ mean that we are very close to machine epsilon ($10^{-7}$).  Another factor is the use of stochastic, gradient-based optimizers. While the Adam optimizer is not stochastic in this case since we are not training with mini-batches of data, it is still true that we are using non-optimal step sizes. Therefore, we hypothesize that the method is getting stuck at errors $\mathcal{O}(10^{-6})$ instead of reaching machine epsilon values $\mathcal{O}(10^{-7})$ in single-precision.
 
\begin{figure}
    \centering
    \includegraphics[width=\textwidth]{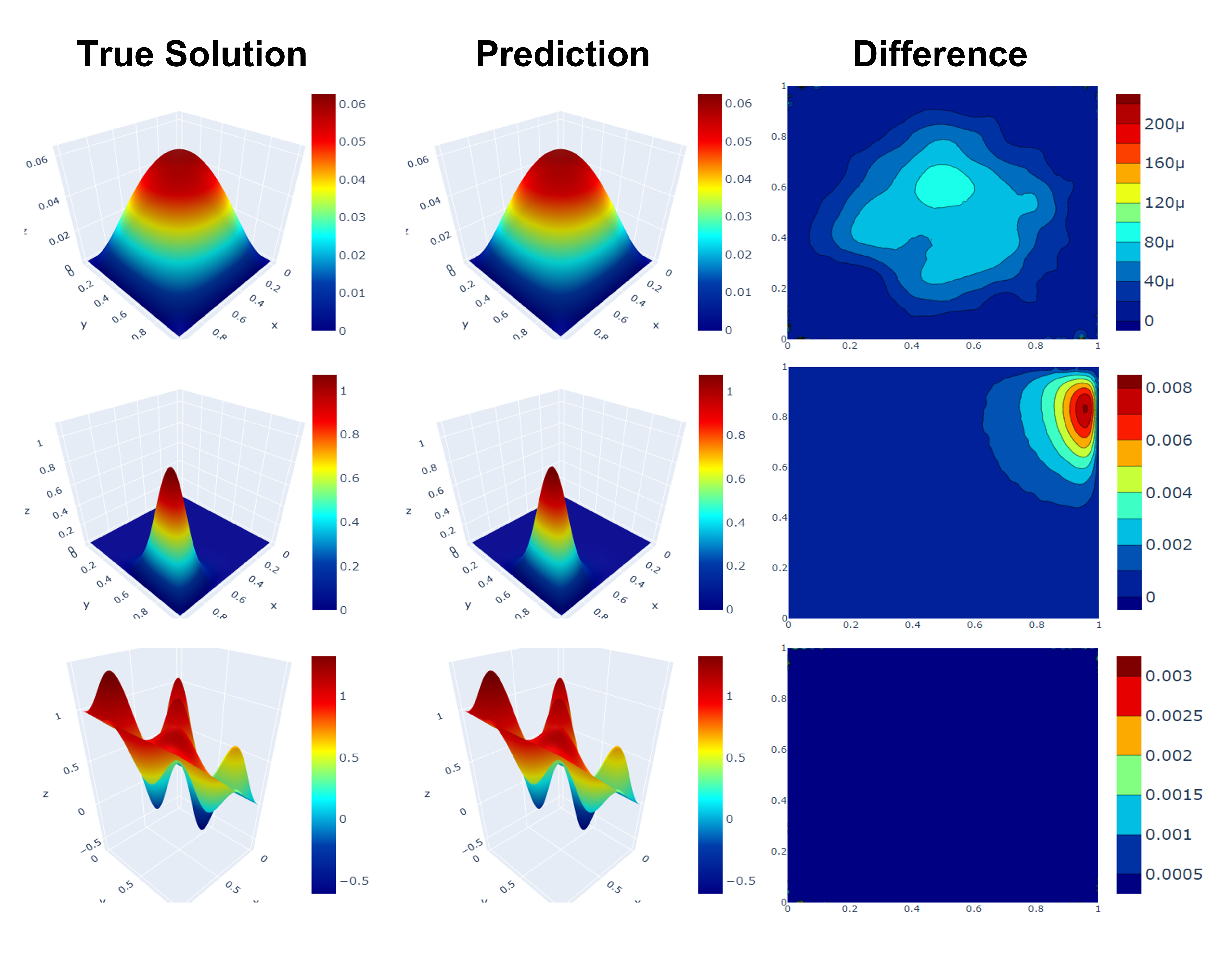}
    \caption{(Top) Bubble function. (Middle) ``Peak'' function. (Bottom) Exponential trigonometric function. From left to right, contour plots of true solution, predicted solution, and absolute difference. All predictions and solutions are on a 128$\times$128 grid. Note that $\mu = 10^{-6}$ in the colorbar for the bubble function difference.}
    \label{fig:elliptic-results}
\end{figure}

\begin{figure}
    \centering
    \includegraphics[width=\textwidth]{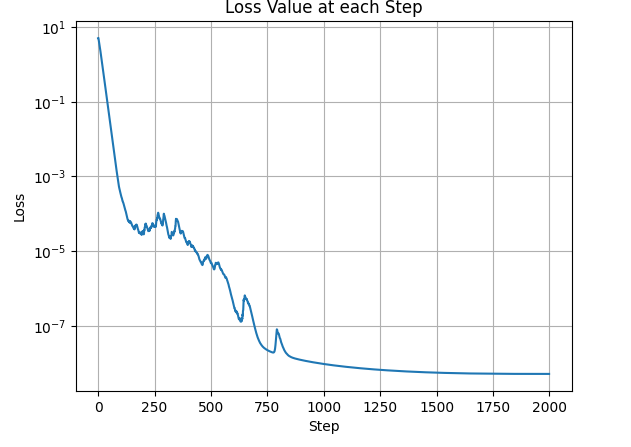}
    \caption{Loss values at each optimization step for the bubble function. Here, we use a network depth of three, a grid size of 128$\times$128, and 2,000 optimization steps.}
    \label{fig:elliptic-loss}
\end{figure}

Figure \ref{fig:elliptic-results} shows the true solution, predicted solution, and absolute difference for the bubble, peak, and exponential-trigonometric cases on a 128$\times$128 grid. This figure shows that for every case, our unsupervised algorithm produces visually indistinguishable solutions from the true solution in each case. To give more insight into our method's performance, we show the loss function for one of the test cases (the bubble function) in Figure~\ref{fig:elliptic-loss}. Here, we observe that our loss function reaches values that are machine epsilon for single precision (on the order of $10^{-7}$). Similar loss function values are observed for the other cases, indicating that we have reached convergence. 

We also want to study the network depth's effect on our predictions' accuracy. Table \ref{tab:results-elliptic-grid-depths} shows the accuracy of our unsupervised neural network predictions for varying grid sizes and U-Net depths. We set the number of optimization steps to 4,000. Here, we see that the depth of the U-Net architectures does not appear to have a significant effect on the accuracy of our predictions. The only notable exceptions are for the bubble and exponential-trigonometric functions on a 128$\times$128 grid with a depth equal to two. This discrepancy may be caused by the network not capturing sufficiently rich features on the finer grid. The fact that we see a decrease in the errors between our predictions and the true solution to those comparable to the finite difference method as we increase the depth to three or greater supports this hypothesis. Note that for coarser grids (i.e., fewer grid points), we do not test higher network depths since the size of the features (the output of each layer) at the coarser grids in such networks would be smaller than the size of the convolutional kernels in the network.

\begin{figure}
    \centering
    \includegraphics[width=\textwidth]{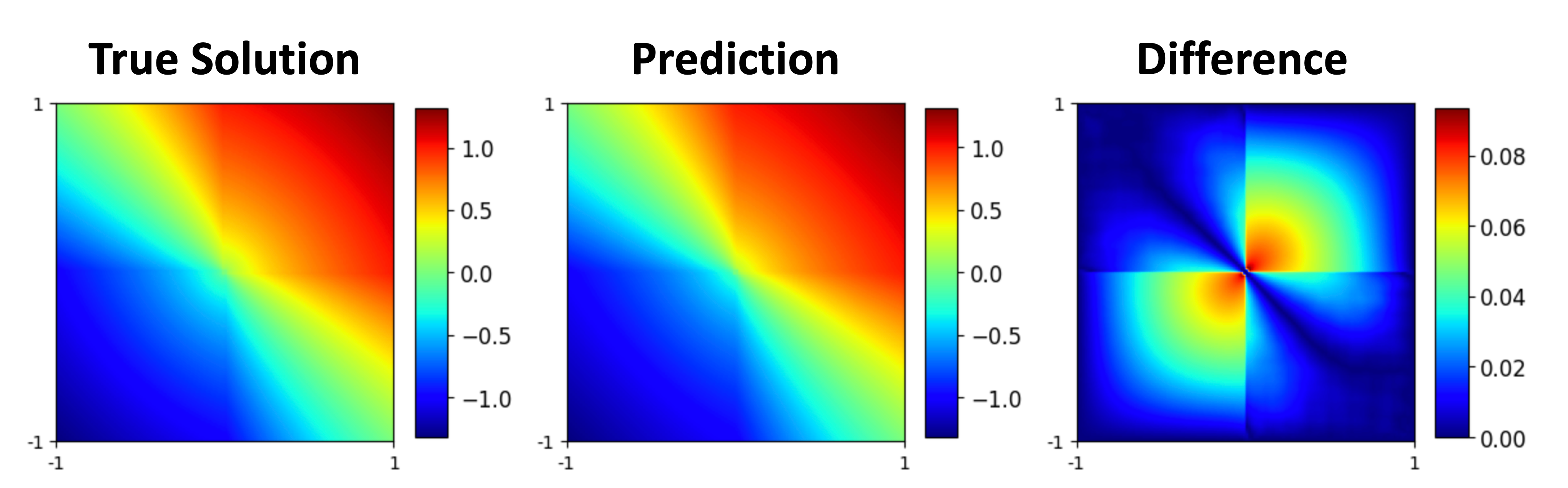}
    \caption{From left to right, contour plots of true solution, predicted solution, and absolute difference for the non-constant diffusion problem.}
    \label{fig:non-const-diff-results}
\end{figure}

The results of Algorithm \ref{alg:alg-elliptic} for the non-constant diffusion problem defined in Section \ref{sec:data-elliptic} are shown in Table \ref{tab:results-non-const-grid-opt-steps}. Here, we see more significant errors than with the constant diffusion problems. Figure \ref{fig:non-const-diff-results} shows the true solution, predicted solution, and absolute difference for the non-constant diffusion problem on a 128$\times$128 grid. This figure shows that the errors are mostly concentrated around the discontinuities in $\kappa$. We do not compare this to the finite difference method because the exact solution exhibits a singularity at the origin. Indeed, the gradient of the exact solution blows up at the point $(0,0)$. 

Because the finite difference solution solves the convex minimization problem (\ref{eqn:opt}), the finite difference approximation's accuracy limits our approach's accuracy. This explains the relatively poor performance in the case of discontinuous diffusion. It is well known that, due to the singularity at the origin, the exact solution to the discontinuous diffusion problem belongs to the Sobolev space $H^{1+\beta}(\Omega$), where $\beta \approx 0.5354$ \cite{riviere2003posteriori}. Hence, the finite difference method, which requires more regularity, performs poorly.

\begin{table}[ht!]
\centering
\bgroup
\def\arraystretch{1.1}
\resizebox{\textwidth}{!}{%
\begin{tabular}{cccc|cc|cc}
\hline
\multirow{2}{*}{Problem} & \multirow{2}{*}{$\mathcal{M}$} & \multicolumn{2}{c|}{$N = 32$} & \multicolumn{2}{c|}{$N = 64$} & \multicolumn{2}{c}{$N = 128$} \\ \cline{3-8} 
 &  & $||u - \hat{u}||_{2,h}$ & $||u - \hat{u}||_{\infty}$ & $||u - \hat{u}||_{2,h}$ & $||u - \hat{u}||_{\infty}$ & $||u - \hat{u}||_{2,h}$ & $||u - \hat{u}||_{\infty}$ \\ \hline
\multirow{5}{*}{Bubble} & 500 & 4.7521e-05 & 5.2953e-04 & 4.8521e-05 & 8.2744e-04 & 4.2407e-04 & 1.6607e-03 \\
 & 1,000 & 2.5330e-05 & 2.4341e-04 & 2.1550e-05 & 3.4875e-04 & 1.3244e-04 & 5.3431e-04 \\
 & 2,000 & 1.4457e-05 & 1.4313e-04 & 9.1955e-06 & 1.4581e-04 & 4.0090e-05 & 3.0874e-04 \\
 & 4,000 & 3.2266e-06 & 3.0962e-05 & 3.1159e-06 & 5.4674e-05 & 7.6852e-06 & 8.5074e-05 \\
 & 8,000 & 2.2661e-06 & 2.9351e-05 & 3.7016e-06 & 4.9231e-05 & 3.7105e-06 & 4.4149e-05 \\ \hline
\multirow{5}{*}{Peak} & 500 & 2.0066e-02 & 1.2398e-01 & 1.0361e-02 & 1.2776e-01 & 3.2611e-02 & 2.8655e-01 \\
 & 1,000 & 2.0603e-02 & 1.1937e-01 & 5.3352e-03 & 3.2162e-02 & 5.0469e-03 & 2.6498e-02 \\
 & 2,000 & 2.0594e-02 & 1.1934e-01 & 5.5620e-03 & 3.2176e-02 & 1.3510e-03 & 8.0407e-03 \\
 & 4,000 & 2.1275e-02 & 1.1932e-01 & 5.6542e-03 & 3.2192e-02 & 1.3851e-03 & 8.0365e-03 \\
 & 8,000 & 2.0616e-02 & 1.1933e-01 & 5.5843e-03 & 3.2196e-02 & 1.4042e-03 & 8.0850e-03 \\ \hline
\multirow{5}{*}{Exp-Trig} & 500 & 2.8224e-03 & 1.8145e-02 & 1.0084e-03 & 1.5877e-02 & 2.6316e-03 & 3.0236e-02 \\
 & 1,000 & 2.4582e-03 & 7.8004e-03 & 6.3506e-04 & 4.8903e-03 & 4.0585e-04 & 1.6215e-02 \\
 & 2,000 & 2.3809e-03 & 7.8190e-03 & 5.9138e-04 & 1.8863e-03 & 1.8116e-04 & 3.6817e-03 \\
 & 4,000 & 2.4604e-03 & 7.8169e-03 & 5.9671e-04 & 2.4966e-03 & 1.5110e-04 & 1.3024e-03 \\
 & 8,000 & 2.3742e-03 & 7.8151e-03 & 5.8227e-04 & 1.8845e-03 & 1.4478e-04 & 5.7032e-04 \\ \hline
\end{tabular}%
}
\egroup
\caption{Accuracy of unsupervised predictions for a varying grid size $N$, number of optimization steps $\mathcal{M}$, and the depth of the U-Net set to three. \label{tab:results-elliptic-grid-opt-steps}}
\end{table}

\begin{table}[ht!]
\centering
\bgroup
\def\arraystretch{1.1}
\resizebox{\textwidth}{!}{%
\begin{tabular}{ccccc|cc|cc}
\hline
\multirow{2}{*}{Problem} & \multirow{2}{*}{$d$} & \multirow{2}{*}{\# Params.} & \multicolumn{2}{c|}{$N = 32$} & \multicolumn{2}{c|}{$N = 64$} & \multicolumn{2}{c}{$N = 128$} \\ \cline{4-9} 
 &  &  & $||u - \hat{u}||_{2,h}$ & $||u - \hat{u}||_{\infty}$ & $||u - \hat{u}||_{2,h}$ & $||u - \hat{u}||_{\infty}$ & $||u - \hat{u}||_{2,h}$ & $||u - \hat{u}||_{\infty}$ \\ \hline
\multirow{4}{*}{Bubble} & 2 & 283,713 & 3.2705e-06 & 3.8412e-05 & 2.8414e-06 & 4.1518e-05 & 5.4252e-04 & 1.9439e-03 \\
 & 3 & 412,225 & 3.2266e-06 & 3.0962e-05 & 3.1159e-06 & 5.4674e-05 & 7.6852e-06 & 8.5074e-05 \\
 & 4 & 541,889 & - & - & 2.6481e-06 & 4.1081e-05 & 3.2062e-06 & 6.0881e-05 \\
 & 5 & 669,249 & - & - & - & - & 2.8731e-06 & 9.5620e-05 \\ \hline
\multirow{4}{*}{Peak} & 2 & 283,713 & 2.1274e-02 & 1.1932e-01 & 5.6116e-03 & 3.2174e-02 & 1.4157e-03 & 7.9525e-03 \\
 & 3 & 412,225 & 2.1275e-02 & 1.1932e-01 & 5.6542e-03 & 3.2192e-02 & 1.3851e-03 & 8.0365e-03 \\
 & 4 & 541,889 & - & - & 5.6673e-03 & 3.2192e-02 & 1.4087e-03 & 8.0848e-03 \\
 & 5 & 669,249 & - & - & - & - & 1.4042e-03 & 8.0918e-03 \\ \hline
\multirow{4}{*}{Exp-Trig} & 2 & 283,713 & 2.4689e-03 & 7.8191e-03 & 5.9526e-04 & 1.8814e-03 & 2.8053e-02 & 7.5356e-02 \\
 & 3 & 412,225 & 2.4604e-03 & 7.8169e-03 & 5.9671e-04 & 2.4966e-03 & 1.5110e-04 & 1.3024e-03 \\
 & 4 & 541,889 & - & - & 5.9509e-04 & 1.8855e-03 & 1.5166e-04 & 1.8679e-03 \\
 & 5 & 669,249 & - & - & - & - & 1.4944e-04 & 1.2124e-03 \\ \hline
\end{tabular}%
}
\egroup
\caption{Accuracy of unsupervised predictions for varying grid sizes $N$ and U-Net depths $d$. We set the number of optimization steps to 4,000.} 
\label{tab:results-elliptic-grid-depths}
\end{table}

\begin{table}[ht!]
\centering
\bgroup
\def\arraystretch{1.2}
\resizebox{0.95\textwidth}{!}{%
\begin{tabular}{ccc|cc|cc}
\hline
\multicolumn{1}{l}{\multirow{2}{*}{Problem}} &
  \multicolumn{2}{c|}{$N = 32$} &
  \multicolumn{2}{c|}{$N = 64$} &
  \multicolumn{2}{c}{$N = 128$} \\ \cline{2-7} 
\multicolumn{1}{l}{} &
 $||u - u_h||_{2,h}$ & $||u - u_h||_{\infty}$ & $||u - u_h||_{2,h}$ & $||u - u_h||_{\infty}$ & $||u - u_h||_{2,h}$ & $||u - u_h||_{\infty}$  \\ \hline
Bubble   & 1.3582e-16 & 2.8449e-16 & 2.4524e-16 & 6.8695e-16 & 8.1134e-16 & 1.9776e-15\\
\hline
Peak    & 1.7597e-02 & 1.1151e-01 & 5.3441e-03& 3.1282e-02 & 1.3988e-03 & 7.9696e-03\\
\hline
Exp-Trig  & 2.2986e-03 & 7.3266e-03 & 5.7262e-04 & 1.8251e-03 & 1.4303e-04 & 4.5588e-04\\
\hline
\end{tabular}%
}
\egroup
\caption{Finite difference errors for selected elliptic problems for comparison.}
\label{tab:fd-results}
\end{table}

\begin{table}[ht!]
\centering
\bgroup
\def\arraystretch{1.1}
\resizebox{\textwidth}{!}{%
\begin{tabular}{cccc|cc|cc}
\hline
\multirow{2}{*}{Problem} & \multirow{2}{*}{$\mathcal{M}$} & \multicolumn{2}{c|}{$N = 32$} & \multicolumn{2}{c|}{$N = 64$} & \multicolumn{2}{c}{$N = 128$} \\ \cline{3-8} 
 &  & $||u - \hat{u}||_{2,h}$ & $||u - \hat{u}||_{\infty}$ & $||u - \hat{u}||_{2,h}$ & $||u - \hat{u}||_{\infty}$ & $||u - \hat{u}||_{2,h}$ & $||u - \hat{u}||_{\infty}$ \\ \hline
\multirow{5}{*}{\begin{tabular}[c]{@{}c@{}}Non. Const. \\ Diff.\end{tabular}} & 500 & 4.6597e-02 & 1.8380e-01 & 8.6890e-02 & 3.7069e-01 & 5.5651e-01 & 2.1105e+00 \\
 & 1,000 & 3.9795e-02 & 1.6035e-01 & 4.9640e-02 & 1.3376e-01 & 2.0766e-01 & 6.4809e-01 \\
 & 2,000 & 3.9718e-02 & 1.5799e-01 & 3.1458e-02 & 1.2455e-01 & 8.7473e-02 & 2.8797e-01 \\
 & 4,000 & 3.9713e-02 & 1.5794e-01 & 3.1224e-02 & 1.2258e-01 & 4.9868e-02 & 1.1281e-01 \\
 & 8,000 & 3.9710e-02 & 1.5800e-01 & 3.1045e-02 & 1.2221e-01 & 3.1002e-02 & 9.3761e-02 \\ \hline
\end{tabular}%
}
\egroup
\caption{Accuracy of unsupervised predictions on the non-constant diffusion problem for a varying grid sizes $N$, number of optimization steps $\mathcal{M}$, and the depth of the U-Net set to three. \label{tab:results-non-const-grid-opt-steps}}
\end{table}

\begin{table}[ht!]
\centering
\bgroup
\def\arraystretch{1.1}
\resizebox{\textwidth}{!}{%
\begin{tabular}{ccccc|cc|cc}
\hline
\multirow{2}{*}{Problem} & \multirow{2}{*}{$d$} & \multirow{2}{*}{\# Params.} & \multicolumn{2}{c|}{$N = 32$} & \multicolumn{2}{c|}{$N = 64$} & \multicolumn{2}{c}{$N = 128$} \\ \cline{4-9} 
 &  &  & $||u - \hat{u}||_{2,h}$ & $||u - \hat{u}||_{\infty}$ & $||u - \hat{u}||_{2,h}$ & $||u - \hat{u}||_{\infty}$ & $||u - \hat{u}||_{2,h}$ & $||u - \hat{u}||_{\infty}$ \\ \hline
\multirow{4}{*}{\begin{tabular}[c]{@{}c@{}}Non. Const. \\ Diff.\end{tabular}} & 2 & 283,713 & 3.9709e-02 & 1.5798e-01 & 3.1294e-02 & 1.2265e-01 & 3.3859e-02 & 1.4377e-01 \\
 & 3 & 412,225 & 3.9713e-02 & 1.5794e-01 & 3.1224e-02 & 1.2258e-01 & 4.9868e-02 & 1.1281e-01 \\
 & 4 & 541,889 & - & - & 3.1110e-02 & 1.2282e-01 & 3.8937e-02 & 1.0907e-01 \\
 & 5 & 669,249 & - & - & - & - & 3.6566e-02 & 1.0511e-01 \\ \hline
\end{tabular}%
}
\egroup
\caption{Accuracy of unsupervised predictions on non-constant diffusion problem for varying grid sizes $N$ and U-Net depths $d$. We set the number of optimization steps to 4,000. \label{tab:results-non-constant-grid-depths}}
\end{table}

\begin{figure}[ht!]
    \centering
    \includegraphics[width=0.95\textwidth]{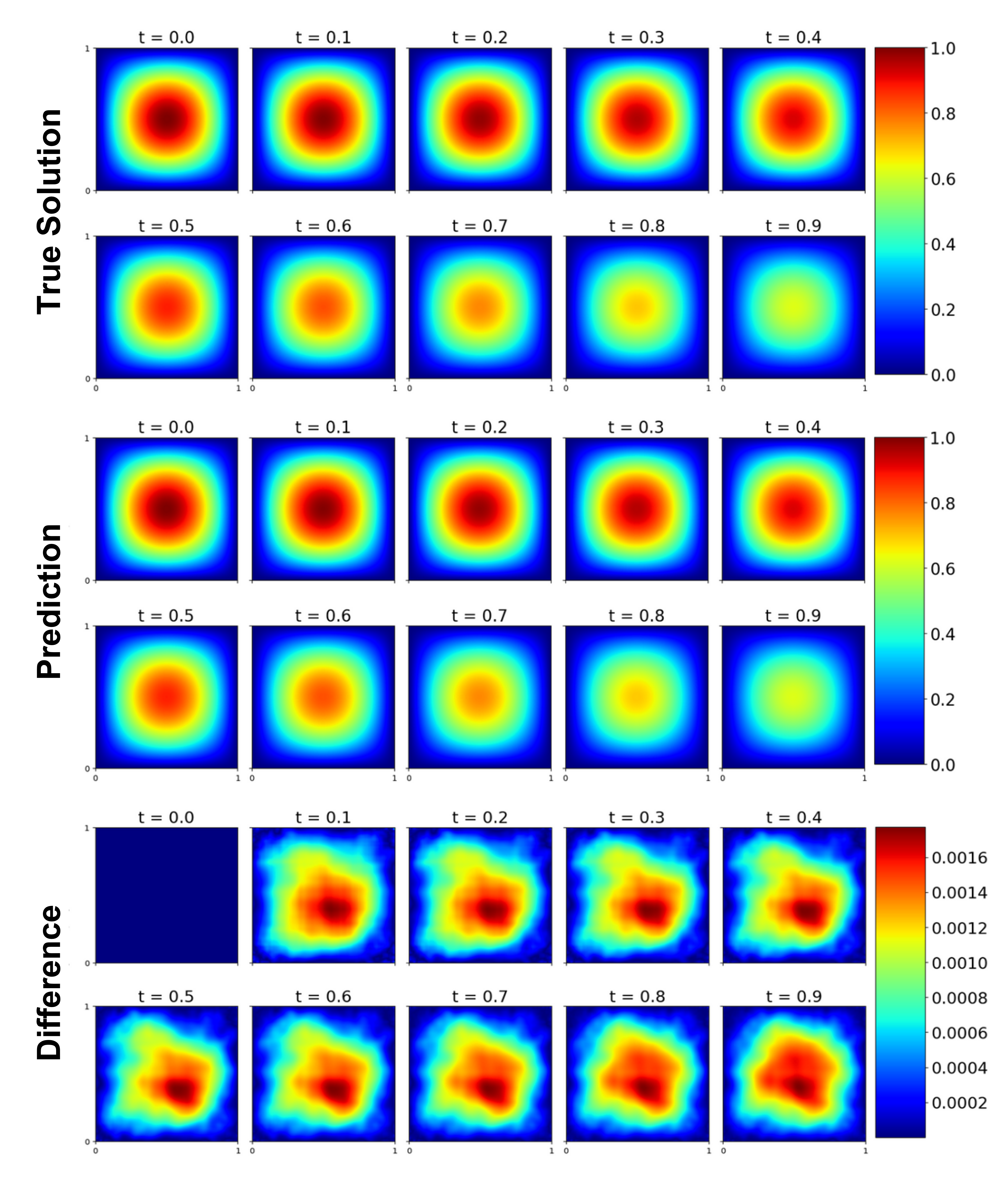}
    \caption{True solution (top), predicted solution (middle) and absolute difference (bottom) for the first ten time steps of the trigonometric problem presented in Section \ref{sec:data} with $n=1$.}
    \label{fig:parabolic-trig-2}
\end{figure}

\begin{figure}[ht!]
    \centering
    \includegraphics[width=0.95\textwidth]{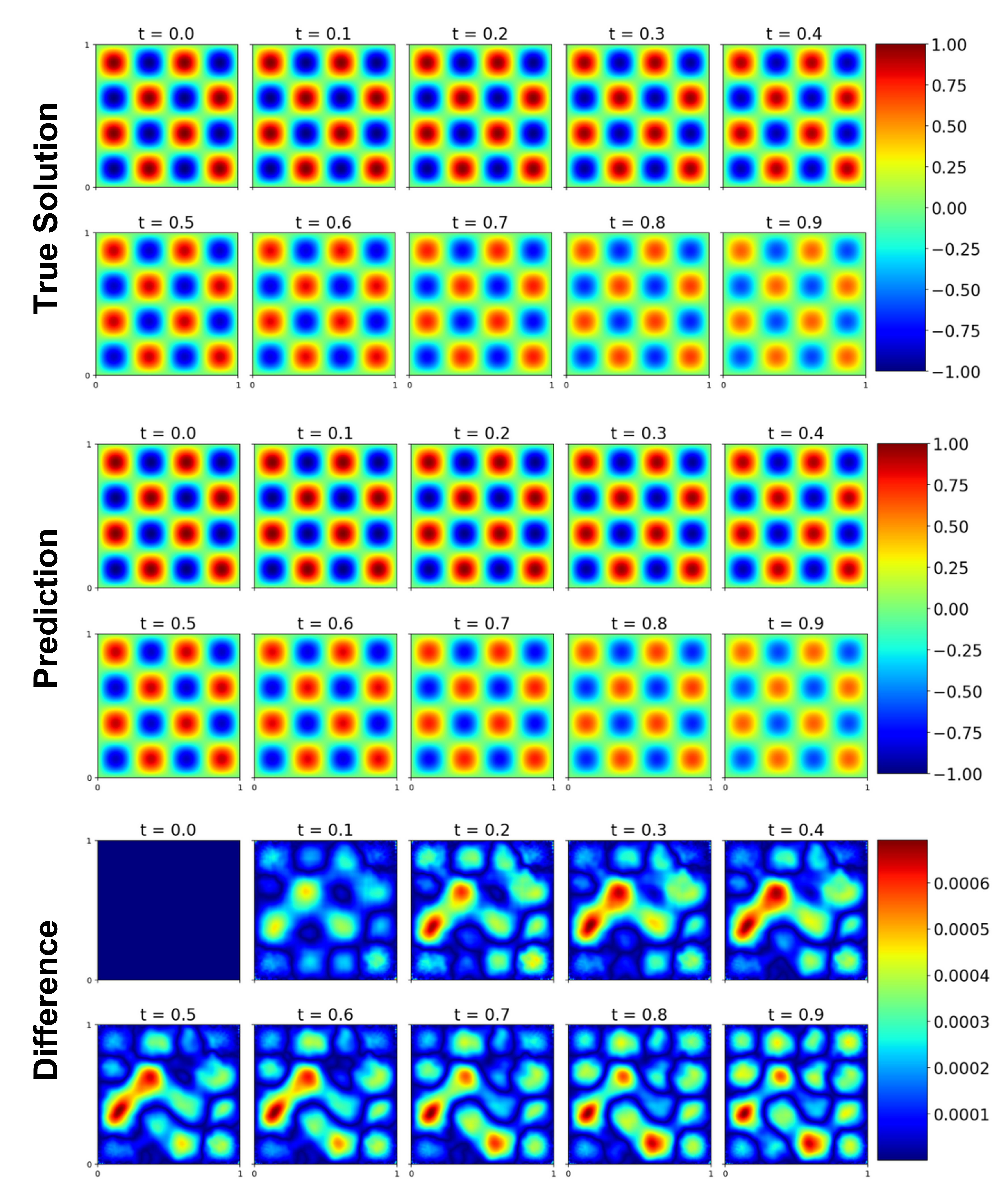}
    \caption{True solution (top), predicted solution (middle) and absolute difference (bottom) for the first ten time steps of the trigonometric problem presented in Section \ref{sec:data} with $n=4$.}
    \label{fig:parabolic-trig-4}
\end{figure}

\begin{figure}[ht!]
    \centering
    \includegraphics[width=0.95\textwidth]{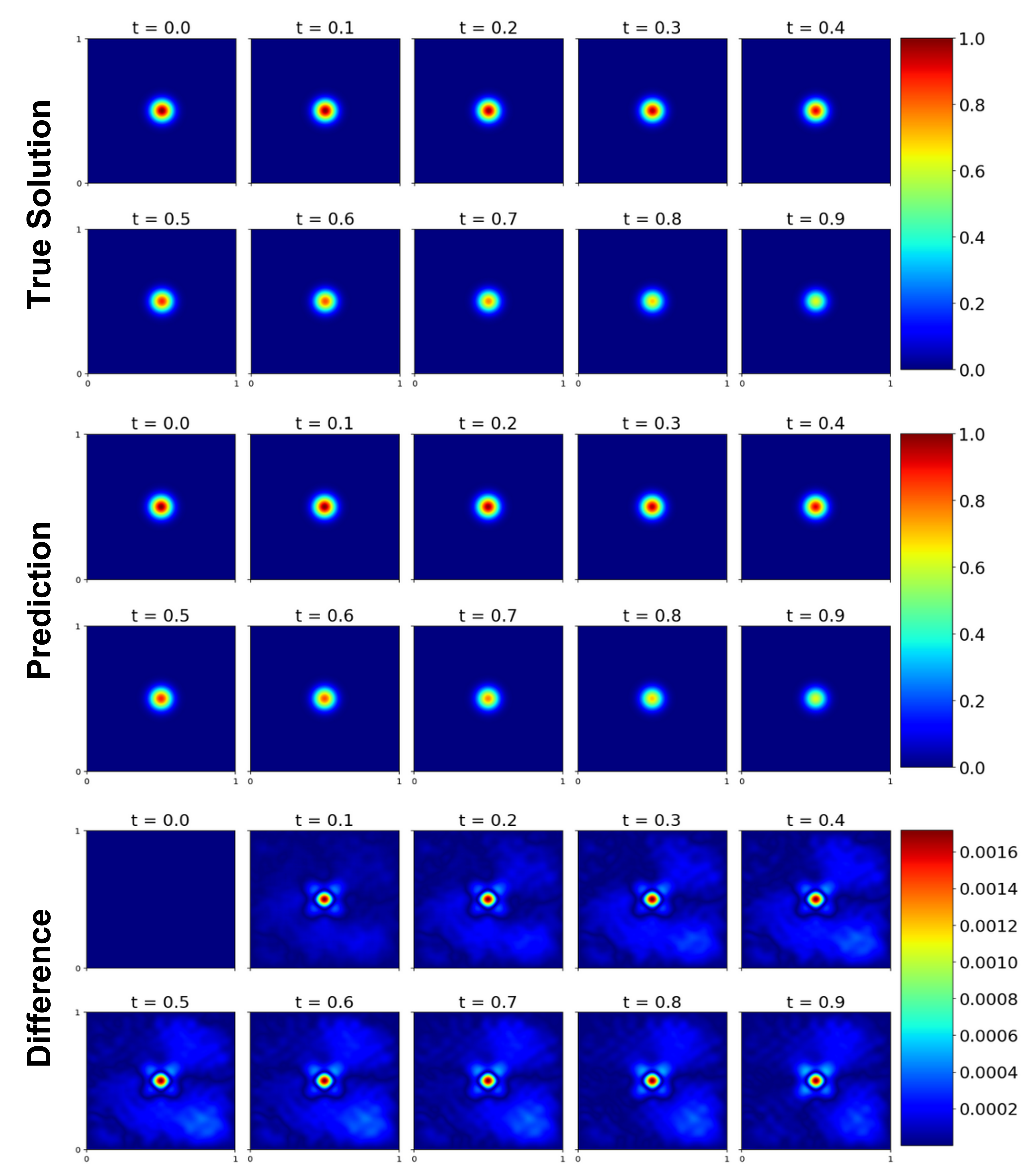}
    \caption{True solution (top), predicted solution (middle) and absolute difference (bottom) for the first ten time steps of the Gaussian problem presented in Section \ref{sec:data}.}
    \label{fig:parabolic-gaussian}
\end{figure}

Using the methods described in Section \ref{sec:methods} and Algorithm \ref{alg:alg-parabolic}, we approximate the finite difference solution to the parabolic problems defined in Section \ref{sec:data-parabolic}. The time step is $\tau = 0.1$. Except for the first time step, we set the number of optimization steps per time step to 250. Because we start with randomly initialized weights in the first time step, we set the number of optimization steps to 1,000 in the first time step. Table \ref{tab:results-parabolic-grid-times} shows the accuracy of our unsupervised predictions for varying grid sizes. For reference, Table \ref{tab:results-fd-parabolic} shows the accuracy of the finite difference method with backward Euler for varying grid sizes. Here, we see that Algorithm \ref{alg:alg-parabolic} achieves comparable accuracy to the finite difference method with backward Euler. However, we do not see the same convergence as with the finite difference approach for the trigonometric functions. The small number of optimization steps can explain this lack of convergence and, again, the weakness of first-order optimizers like Adam. Figures \ref{fig:parabolic-trig-2} through \ref{fig:parabolic-gaussian} show the true solution, predicted solution, and absolute difference for the first ten time steps of each problem. These figures show that our unsupervised approach produces visually accurate solutions. Finally, Table \ref{tab:results-parabolic-grid-activations} displays the errors for several choices of activation functions at time $t=0.5$. The errors are similar for ReLU, Tanh, Swish \cite{ramachandran2017searching}, and identity activation functions. These results indicate that linear activation functions (i.e., the identity) are sufficient for learning solutions to time-dependent problems. 

\begin{table}[ht!]
\centering
\bgroup
\def\arraystretch{1.2}
\resizebox{\textwidth}{!}{%
\begin{tabular}{cccc|cc|cc}
\hline
\multirow{2}{*}{Problem} & \multirow{2}{*}{$t$} & \multicolumn{2}{c|}{$N = 32$} & \multicolumn{2}{c|}{$N = 64$} & \multicolumn{2}{c}{$N = 128$} \\ \cline{3-8} 
 &  & $||u - \hat{u}||_{2,h}$ & $||u - \hat{u}||_{\infty}$ & $||u - \hat{u}||_{2,h}$ & $||u - \hat{u}||_{\infty}$ & $||u - \hat{u}||_{2,h}$ & $||u - \hat{u}||_{\infty}$ \\ \hline
\multirow{4}{*}{Trig. ($n = 1$)} & 0.5 & 7.6825e-04 & 1.5388e-03 & 1.0661e-03 & 2.1703e-03 & 1.1562e-03 & 2.4109e-03 \\
 & 1.0 & 5.2059e-04 & 1.0459e-03 & 7.0987e-04 & 1.4189e-03 & 7.7541e-04 & 1.5364e-03 \\
 & 2.5 & 6.1710e-04 & 1.2354e-03 & 8.6825e-04 & 1.7483e-03 & 9.4440e-04 & 1.9382e-03 \\
 & 5.0 & 1.6629e-04 & 3.3075e-04 & 2.5692e-04 & 5.2649e-04 & 3.0442e-04 & 6.2141e-04 \\ \hline
\multirow{4}{*}{Trig. ($n = 4$)} & 0.5 & 6.0011e-03 & 1.1977e-02 & 1.3953e-03 & 2.9254e-03 & 3.2371e-04 & 9.0033e-04 \\
 & 1.0 & 3.6984e-03 & 7.3841e-03 & 8.5877e-04 & 1.8117e-03 & 1.9062e-04 & 5.3841e-04 \\
 & 2.5 & 5.4592e-03 & 1.0897e-02 & 1.3062e-03 & 3.1186e-03 & 8.4891e-04 & 2.8031e-03 \\
 & 5.0 & 1.9352e-03 & 3.8744e-03 & 4.8639e-04 & 1.3043e-03 & 9.8322e-04 & 2.5854e-03 \\ \hline
\multirow{4}{*}{Gaussian} & 0.5 & 3.0932e-03 & 3.8826e-02 & 7.1997e-04 & 1.0341e-02 & 1.8902e-04 & 2.5298e-03 \\
 & 1.0 & 1.9069e-03 & 2.3921e-02 & 4.4389e-04 & 6.3892e-03 & 1.3421e-04 & 1.4701e-03 \\
 & 2.5 & 2.8197e-03 & 3.5385e-02 & 6.5362e-04 & 9.5078e-03 & 2.7103e-04 & 2.3168e-03 \\
 & 5.0 & 1.0004e-03 & 1.2579e-02 & 2.8684e-04 & 2.6686e-03 & 3.1119e-04 & 9.0614e-04 \\ \hline
\end{tabular}%
}
\egroup
\caption{Accuracy of unsupervised predictions for varying grid sizes and time steps for parabolic problems. The depth of the network is set to three, the number of optimization iterations at each time step to 250, and the time step to 0.1. \label{tab:results-parabolic-grid-times}}
\end{table}

\begin{table}[ht!]
\centering
\bgroup
\def\arraystretch{1.2}
\resizebox{\textwidth}{!}{%
\begin{tabular}{cccc|cc|cc}
\hline
\multirow{2}{*}{Problem} &
  \multirow{2}{*}{$t$} &
  \multicolumn{2}{c|}{$N = 32$} &
  \multicolumn{2}{c|}{$N = 64$} &
  \multicolumn{2}{c}{$N = 128$} \\ \cline{3-8} 
 &
   &
  $||u - \hat{u}||_{2,h}$ &
  $||u - \hat{u}||_{\infty}$ &
  $||u - \hat{u}||_{2,h}$ &
  $||u - \hat{u}||_{\infty}$ &
  $||u - \hat{u}||_{2,h}$ &
  $||u - \hat{u}||_{\infty}$ \\ \hline
\multirow{4}{*}{Trig. ($n = 1$)} & 0.5 & 7.6693e-04 & 1.5299e-03 & 1.0560e-03 & 2.1106e-03 & 1.1255e-03 & 2.2507e-03 \\
                                 & 1.0 & 5.2033e-04 & 1.0380e-03 & 7.0797e-04 & 1.4151e-03 & 7.5314e-04 & 1.5060e-03 \\
                                 & 2.5 & 6.1623e-04 & 1.2293e-03 & 8.6455e-04 & 1.7280e-03 & 9.2434e-04 & 1.8484e-03 \\
                                 & 5.0 & 1.5534e-04 & 3.0988e-04 & 2.3116e-04 & 4.6204e-04 & 2.4942e-04 & 4.9877e-04 \\ \hline
\multirow{4}{*}{Trig. ($n = 4$)} & 0.5 & 5.9961e-03 & 1.1961e-02 & 1.3890e-03 & 2.7763e-03 & 2.8791e-04 & 5.7573e-04 \\
                                 & 1.0 & 3.7021e-03 & 7.3852e-03 & 8.5647e-04 & 1.7119e-03 & 1.7638e-04 & 3.5271e-04 \\
                                 & 2.5 & 5.4540e-03 & 1.0880e-02 & 1.2656e-03 & 2.5296e-03 & 2.6449e-04 & 5.2889e-04 \\
                                 & 5.0 & 2.8215e-03 & 5.6286e-03 & 6.5597e-04 & 1.3111e-03 & 1.3834e-04 & 2.7664e-04 \\ \hline
\multirow{4}{*}{Gaussian}        & 0.5 & 3.0933e-03 & 3.8814e-02 & 7.1809e-04 & 1.0412e-02 & 1.8424e-04 & 2.4368e-03 \\
                                 & 1.0 & 1.9068e-03 & 2.3920e-02 & 4.4286e-04 & 6.4085e-03 & 1.1490e-04 & 1.4920e-03 \\
                                 & 2.5 & 2.8194e-03 & 3.5387e-02 & 6.5422e-04 & 9.5075e-03 & 1.6586e-04 & 2.2389e-03 \\
                                 & 5.0 & 1.4619e-03 & 1.8355e-02 & 3.3909e-04 & 4.9403e-03 & 8.4882e-05 & 1.1716e-03 \\ \hline
\end{tabular}%
}
\egroup
\caption{Finite differences with backward Euler errors on selected parabolic problems for comparison. Here, the time step is 0.1.}
\label{tab:results-fd-parabolic}
\end{table}

\begin{table}[ht!]
\centering
\bgroup
\def\arraystretch{1.2}
\resizebox{\textwidth}{!}{%
\begin{tabular}{cccc|cc|cc}
\hline
\multirow{2}{*}{Problem} & \multirow{2}{*}{Activation} & \multicolumn{2}{c|}{$N = 32$} & \multicolumn{2}{c|}{$N = 64$} & \multicolumn{2}{c}{$N = 128$} \\ \cline{3-8} 
 &  & $||u - \hat{u}||_{2,h}$ & $||u - \hat{u}||_{\infty}$ & $||u - \hat{u}||_{2,h}$ & $||u - \hat{u}||_{\infty}$ & $||u - \hat{u}||_{2,h}$ & $||u - \hat{u}||_{\infty}$ \\ \hline
\multirow{4}{*}{Trig. ($n = 1$)} & ReLU & 7.6775e-04 & 1.5420e-03 & 1.0746e-03 & 2.1598e-03 & 1.2400e-03 & 2.5946e-03 \\
 & Tanh & 7.6743e-04 & 1.5391e-03 & 1.0590e-03 & 2.1269e-03 & 1.1478e-03 & 2.3299e-03 \\
 & Swish & 7.6905e-04 & 1.5538e-03 & 1.0623e-03 & 2.0923e-03 & 1.1637e-03 & 2.4307e-03 \\
 & Identity & 7.6825e-04 & 1.5388e-03 & 1.0661e-03 & 2.1703e-03 & 1.1562e-03 & 2.4109e-03 \\ \hline
\multirow{4}{*}{Trig. ($n = 4$)} & ReLU & 5.9967e-03 & 1.1968e-02 & 1.5607e-03 & 4.1957e-03 & 9.2973e-04 & 2.7800e-03 \\
 & Tanh & 5.9970e-03 & 1.1968e-02 & 1.3906e-03 & 2.8266e-03 & 3.1624e-04 & 8.8251e-04 \\
 & Swish & 6.0061e-03 & 1.1990e-02 & 1.6013e-03 & 3.9862e-03 & 4.7380e-04 & 1.4040e-03 \\
 & Identity & 6.0011e-03 & 1.1977e-02 & 1.3953e-03 & 2.9254e-03 & 3.2371e-04 & 9.0033e-04 \\ \hline
\multirow{4}{*}{Gaussian} & ReLU & 3.0932e-03 & 3.8828e-02 & 7.3665e-04 & 1.0873e-02 & 3.8312e-04 & 3.3270e-03 \\
 & Tanh & 3.0930e-03 & 3.8808e-02 & 7.1797e-04 & 1.0419e-02 & 1.9917e-04 & 2.4221e-03 \\
 & Swish & 3.0933e-03 & 3.8830e-02 & 7.4640e-04 & 1.0099e-02 & 6.0038e-04 & 4.4619e-03 \\
 & Identity & 3.0932e-03 & 3.8826e-02 & 7.1997e-04 & 1.0341e-02 & 1.8902e-04 & 2.5298e-03 \\ \hline
\end{tabular}%
}
\egroup
\caption{Accuracy of unsupervised predictions for varying grid sizes and activation functions at time 0.5. The depth of the network is set to three, the number of optimization steps to 250, and the time step is 0.1.}
\label{tab:results-parabolic-grid-activations}
\end{table}

Algorithm \ref{alg:alg-elliptic} may benefit from transfer learning (i.e., reusing weights from previous problems) with a supervised counterpart that is trained on a large labeled dataset. We randomly initialize the neural network weights in Algorithms \ref{alg:alg-elliptic} and \ref{alg:alg-parabolic} (first time step only). This random initialization may result in predictions with high errors at the beginning of training. Without optimal step sizes in our optimizer, these poor solutions may result in slow convergence towards an optimal solution. Using transfer learning with a pre-trained network that is trained in a supervised setting (i.e., with labeled training data), the initial predictions from our algorithms may be already close to optimal, resulting in faster convergence. One could view the use of these pre-trained networks as a sort of preconditioner for our neural network-based approach. Indeed, we see the benefits of transfer learning with Algorithm \ref{alg:alg-parabolic} after the first time step. Instead of reinitializing the weights for the subsequent time steps, we resuse the weights from the previous solution. That allows us to use a small number of optimization steps (i.e., $\mathcal{M} = 250$) at each time for parabolic problems.  

It is important to note that our methods for estimating the finite difference solutions do not explicitly construct the finite difference matrix. Instead, we use our neural networks to map the source term $f$ to the approximate finite difference solution. In this sense, we are learning the inverse mapping of the finite difference matrix. From a numerical point of view, this provides the advantage of not having to construct or store a finite difference matrix. Instead, we implement the required stencil(s) for our problem in the loss function and apply them appropriately. Only having to implement the stencils and not construct a finite difference matrix is also an advantage from an implementation point-of-view. 

The justification for the lack of non-linear activation functions in our network is that the relationship between the source term $f$ and the finite differences solution $u_h$ is given by a system of linear equations of the form $Au_h = f$. In other words, we know a priori that our goal is to learn a linear relationship between $f$ and our estimated solution. In many machine learning applications, the exact nature of the input and output relationship is unknown and assumed to be highly non-linear. Hence, including non-linear activation functions results in a network that learns a non-linear relationship between inputs and outputs. Using the identity function as an activation puts us outside the scope of the universal approximation theorem \cite{horni1989multilayer, cybenko1989approximation, funahashi1989approximate}. However, in our case, we do not need our neural network to be a universal approximator. The role of the network is to learn a linear relationship for a single example. Inputs other than $f$ in Algorithms \ref{alg:alg-elliptic} and \ref{alg:alg-parabolic}, like random noise or constants, would break our assumption that the relationship between the network input and output is linear. Hence, a much larger (i.e., more parameters) non-linear network would be needed to learn such a relationship. 

We utilize the PocketNet approach proposed by \cite{celaya2022pocketnet} in our proposed algorithms. This approach takes advantage of the similarity between the U-Net architecture and geometric multigrid methods to drastically reduce the number of parameters, while maintaining the same accuracy as conventional CNNs for medical imaging and scientific machine learning tasks \cite{zhang2024efficient, celaya2023mgnets, celaya2024inversion}. Additionally, we replace transposed convolution with bilinear upsampling. We find that these changes save time and memory and yield the same accuracy that we see using conventional CNNs (i.e., doubling the number of channels at every depth). These results indicate that smaller neural networks (in terms of parameters) can achieve high accuracy for scientific machine learning tasks.

We see in Tables \ref{tab:results-elliptic-grid-depths} and \ref{tab:results-non-constant-grid-depths} that the depth of the U-Net architecture does not generally have a significant effect on our results. This indicates that, regardless of depth, the U-Net architecture is sufficiently expressive to learn a mapping from the right hand side $f$ to an approximation of the finite difference solution $u_h$. However, non-U-shaped architectures like the HRNet may also produce similar or improved results \cite{hrnet}. Additionally, the use of residual or dense connections within the convolutions of our architecture may also be beneficial \cite{resnet, densenet}. Such block designs have been shown to speed up convergence to lower loss values for neural networks \cite{li2018visualizing}. Finally, modifying our existing architecture by adding deep supervision could also speed up convergence to lower loss values in fewer iterations \cite{pmlr-v38-lee15a, li2022comprehensive}.

The use of the weighting parameter $\alpha$ in (\ref{eqn:loss}) and (\ref{eqn:td-loss}) is necessary to enforce the given boundary conditions. We arrived at our proposed values of $\alpha$ via a grid search over a range of possible values. However, finding optimal values of weighting parameters like $\alpha$ is an open question \cite{cuomo2022scientific}. Strongly enforcing the Dirichlet boundary conditions by modifying the network output to the prescribed values on the boundary is another approach that has been shown to be effective from an experimental and theoretical perspective \cite{berrone2023enforcing, Zhao2023, zeinhofer2023unified}. However, both approaches of weakly and strongly enforcing boundary conditions in neural network-based methods are popular and have their own strengths and weaknesses \cite{cuomo2022scientific}. In our case, strongly enforcing boundary conditions does not have a significant impact on the accuracy of our predicted solutions. Indeed, even with our weakly enforced approach, the boundary is not a significant source of error.

\section{Conclusions}
The results presented above show the effectiveness of our proposed unsupervised approaches for estimating the finite difference solution to elliptic and parabolic problems. Unlike classical PINNs, our approach is influenced by numerical PDEs (i.e., the finite difference method), resulting in more explainable solutions. Additionally, unlike other CNN-based approaches, we define the method for elliptic PDEs with non-constant diffusion coefficients and extend it to time-dependent problems. Finally, we use small linear CNNs, making our method computationally efficient.

Our approach could benefit finite difference solvers by producing better initial guesses and/or acting as a preconditioner. With a few iterations, Algorithm \ref{alg:alg-elliptic} can produce good initial guesses for iterative solvers, thereby reducing the number of iterations required to solve the linear system resulting from (\ref{eqn:fd-scheme}) or (\ref{eqn:non-const-fd}). This same idea can also apply to time-dependent problems, but with initial guesses being produced at each time step. Additionally, because we employ identity for activation functions, our neural networks are linear. Hence, it may be possible to represent a pretrained architecture as a matrix. This resulting matrix could then be used as a preconditioner for finite difference solvers. This use of preconditioners and further testing on other kinds of PDEs, like convection-diffusion, coupled, and nonlinear problems, will be the object of future work. 

\section{Acknowledgements}
The Department of Defense supports Adrian Celaya through the National Defense Science \& Engineering Graduate Fellowship Program. Keegan Kirk is supported by the Natural Sciences and Engineering Research Council of Canada through the Postdoctoral Fellowship Program (PDF-568008). David Fuentes is partially supported by R21CA249373. Beatrice Riviere is partially supported by NSF-DMS2111459. This research was partially supported by the Tumor Measurement Initiative through the MD Anderson Strategic Research Initiative Development (STRIDE), NSF-2111147, and NSF-2111459.

\bibliographystyle{ieeetr}
\bibliography{sources}
\end{document}